\documentclass[a4paper, 10pt]{article}

\usepackage{amsmath,amssymb}
\usepackage{amsthm}
\usepackage[singlelinecheck=off]{caption}
\usepackage{microtype}

\usepackage{subfig}
\usepackage{xcolor}

\usepackage{graphicx}
\usepackage{algorithm,algpseudocode}
\usepackage{mathrsfs}
\usepackage{epstopdf}
\usepackage{float}
\usepackage{url}
\usepackage{accents}

\newcommand{\x}{\mathbf{x}}

\title{Solving for high dimensional committor functions using
  artificial neural networks\thanks{ The work of Y.K. and L.Y. is
    supported in part by the National Science Foundation under award
    DMS-1521830 and the U.S. Department of Energy’s Advanced
    Scientific Computing Research program under award
    DE-FC02-13ER26134/DE-SC0009409.  The work of J.L. is supported in
    part by the National Science Foundation under award
    DMS-1454939. The collaboration is also supported by the National
    Science Foundation Research Networks in Mathematical Sciences
    KI-Net under grant DMS-1107444 and DMS-1107465.}}

\author{Yuehaw Khoo\thanks{Department of Mathematics, Stanford University, Stanford, CA 94305, USA
		(\texttt{ykhoo@stanford.edu}).}
	\and Jianfeng Lu\thanks{Department of Mathematics, Department of Chemistry and Department of
		Physics, Duke University, Durham, NC 27708, USA (\texttt{jianfeng@math.duke.edu}).}
	\and Lexing Ying\thanks{Department of Mathematics and ICME, Stanford University, Stanford, CA
		94305, USA (\texttt{lexing@stanford.edu}).}  }

\begin{document}
	\maketitle
	
\begin{abstract}
  In this note we propose a method based on artificial neural network
  to study the transition between states governed by stochastic
  processes. In particular, we aim for numerical schemes for the
  \emph{committor function}, the central object of transition path
  theory, which satisfies a high-dimensional Fokker-Planck
  equation. By working with the variational formulation of such
  partial differential equation and parameterizing the committor
  function in terms of a neural network, approximations can be
  obtained via optimizing the neural network weights using stochastic
  algorithms. The numerical examples show that moderate accuracy can
  be achieved for high-dimensional problems.
\end{abstract}

\section{Introduction}
In this paper, we study the transition between two states described by the overdamped Langevin process
\begin{equation}
\label{langevin}
d\mathbf{X}_t = -\nabla U(\mathbf{X}_t) dt + \sqrt{2 \beta^{-1}} d\mathbf{W}_t
\end{equation}
where $\mathbf{X}_t \in \Omega \subset \mathbb{R}^d$, $U:\mathbb{R}^d \rightarrow \mathbb{R}$, $\beta = 1/ T$, and $\mathbf{W}_t$ is a $d$-dimensional Wiener process, using the transition path theory \cite{weinan2006towards,EVa:10, LuNolen:15}. The central object in the transition path theory is the \emph{committor function}. Let $\tau_D$ be the first hitting time for region $D \subset \Omega$.  For two disjoint regions $A, B \subset \Omega$, the committor function is defined as
\begin{equation}
q(\x) = \mathbb{P}(\tau_B < \tau_A \mid \mathbf{X}_0 = \x),
\end{equation}
which is the probability of hitting region $B$ before region $A$ with
the stochastic process \eqref{langevin} starting at $\x$. The
committor function $q(\x)$ provides useful statistical description on
properties such as density and probability current of reaction
trajectory. However, obtaining the committor function $q(\x)$ can be a
formiddable task, as it involves solving a high-dimensional
Fokker-Planck equation
\begin{equation}
\label{BK}
0 = -\beta^{-1} \Delta q(\x)+ \nabla U(\x) \cdot \nabla q(\x)\ \ \text{in}\ \Omega \setminus A\cup B, \quad q(\x)\vert_{\partial A} = 0, \quad q(\x)\vert_{ \partial B} = 1.
\end{equation}
The high dimensional nature of \eqref{BK} renders obtaining $q(\x)$
via finite-element-type methods intractable. On the other hand, since
the transition paths are often localized to a quasi one-dimensional
\emph{reaction tube}, the region of interest is rather small compare
to $\Omega$. Under this approximation, the finite temperature string
method \cite{ERenVa:05, VaVe:09} is developed to simultaneously find
the best ``tube'' and the corresponding committor function.  Another
approach is based on explicit dimension reduction using e.g., the
leading eigenfunctions of the generator
$-\beta^{-1} \Delta + \nabla U(\x) \cdot \nabla$ by means of diffusion
maps \cite{coifman2006diffusion,coifman2008diffusion} to approximate
$q(\x)$. More recently, a direct point cloud discretization for the
Fokker-Planck equation \cite{lai2017point} has been also considered.

In recent years, the artificial neural-network (NN) has shown great
success in representing high-dimensional probablity distributions or
classifiers in a variety of machine learning tasks
\cite{hinton2006reducing,lecun2015deep,schmidhuber2015deep}.  
Motivated by those recent success, in this note we use an NN to
provide a low-dimensional parameterization of the committor function
\begin{equation}
q(\x) \rightarrow q_\mathbf{\theta}(\x)
\end{equation}
where $\mathbf{\theta}\in \mathbb{R}^p$ is the parameter vector of the
NN.  $q_\theta(\x)$ is then obtained by minimizing the variational
formulation of \eqref{BK} over $\theta$ as a nonlinear Galerkin
method. In this way, we switch from solving a partial differential
equation to solving an optimization problem. Optimizing the cost in
the variational formulation of \eqref{BK} using gradient-descent-based
method involves computing an integral with respect to the equilibrium
measure. For doing such integration, we use a Monte-Carlo method where
the samples are generated according to the stochastic process
\eqref{langevin}. An NN parameterized committor function can
potentially be used to guide further sampling in transition regions
wherein sample density is low (in similar spirit as \cite{zhang2017reinforced} where an NN parameterized force field is used to guide sampling in the application of molecular dynamics).

This paper is organized as the following. In Section \ref{section:architecture}, we design an NN tailored for solving for the committor function. In Section \ref{section:numerical}, we demonstrate the success of the proposed method in a few examples. In Section \ref{section:conclusion}, we conclude the note. Before moving on, in the next subsection we survey related methods for solving \eqref{BK}.

\subsection{Previous approaches}
A popular way to solve \eqref{BK} is to discretize the generator
$-\beta^{-1} \Delta + \nabla U(\x) \cdot \nabla$ of the overdamped
Langevin process \eqref{langevin} on the sampled points using
diffusion map \cite{coifman2006diffusion,coifman2008diffusion}. The
lower eigenmodes of such discretized operator in principle can provide
a low-dimensional (nonlinear) reparameterization of the committor
function. If the transition trajectories lie on a low dimensional
manifold, it is possible to discretize the generator accurately via
sampling. More precisely, with $N$ samples, let
$K\in \mathbb{R}^{N\times N}$, the diagonal matrix
$D \in \mathbb{R}^{N\times N}$, the discretized generator
$L\in \mathbb{R}^{N\times N}$ be defined as
\begin{equation}
K(\x^i, \x^j) = \frac{\exp(-\vert \x^i-\x^j\vert^2/2 \epsilon^2)}{\sqrt{\exp(-\beta U(\x^i)) \exp(-\beta U(\x^j))}} ,\quad D(\x^i) = \sum_{j}  K(\x^i, \x^j),\quad L =  D^{-1} K - I
\end{equation}
respectively. Let $q_A, q_B, q_{\Omega\setminus A\cup B}$ be vectors corresponding to committor function values on the points belong to regions $A,B, \Omega\setminus A\cup B$, the committor function $q$ satisfies
\begin{equation}
\label{harmonic extension}
L(\Omega\setminus A\cup B,\Omega \setminus A\cup B) q_{\Omega \setminus A\cup B} = - L(\Omega\setminus A\cup B,B) q_B.
\end{equation}
In the presence of a spectral gap the eigenmodes of $L(\Omega\setminus A\cup B,\Omega \setminus A\cup B)$ provides reduced coordinates for the reaction tube, thus solving \eqref{harmonic extension} for $q$ can be seen as expanding $q$ using the reduced coordinates. However, discretizing the generator using diffusion map may suffer from low order of convergence. Therefore \cite{lai2017point} improves upon diffusion map by explicitly constructing the tangent plane of each point in the sampled point cloud and discretizing the generator in each of the tangent plane.

On the other hand, recent years have seen usage of machine learning techniques in solving high-dimensional partial differential equations. The success of \cite{carleo2017solving} where an NN parameterized spin wavefunction is used as an ansatz for solving the many-body Schr\"{o}dinger equation motivates us to consider solving \eqref{BK} using an NN as well. Our work is also similar to the methods in \cite{lagaris1998artificial, sirignano2017dgm,berg2017unified,weinandeep} for solving partial differential equations. \cite{lagaris1998artificial,berg2017unified} demonstrate success of NN-based method for solving boundary values problem 
\begin{equation}
L u = f,\ \x\in\Omega, \quad Bu = g,\ \x\in\partial \Omega
\end{equation}
by  assuming 
\begin{equation}
u(\x) \approx a(\x) + n(\x) b(\x)
\end{equation}
where $b(\x)=0$ on $\partial \Omega$, $a(\x)$ is a smooth function that satisfies the boundary conditions on $\partial \Omega$, and $n(\x)$ is an NN-parameterized function. Then $u$ is found from solving
\begin{equation}
\label{L2 min}
\inf_u \| Lu-f \|_{L_2}^2,
\end{equation}
leading to an optimization problem over the NN parameters. The
improvement of \cite{berg2017unified} over
\cite{lagaris1998artificial} is that $a(\x)$ and $b(\x)$ are also
learned as a neural-network separately from $n(\x)$, whereas in
\cite{lagaris1998artificial} they are obtained via explicit
construction. Such methods remove the need of specifying basis for
discretizing $u$ therefore can complement Galerkin-type method. While
these methods obtain rather impressive results in low-dimension, their
performance in high dimension is unexplored, which is in fact the most
interesting regime.  In a very recent work \cite{sirignano2017dgm}, a
neural-network is used to parameterize the solution to a
high-dimensional parabolic equation. Although such setting is similar
the one we consider, in our case the boundary conditions might
result singularities in $u$, making it more difficult to be
approximated using an NN, which will be address in Section~\ref{section:architecture}. 
Moreover, since \cite{lagaris1998artificial,
  sirignano2017dgm,berg2017unified} work with the strong form of a
partial differential equation, the computational cost can be high as
the second order derivative of $u(\x)$ is needed, whereas our approach
is based on the variational formulation of the PDE. Although using an NN in solving the variational formulation of a PDE \cite{weinandeep} has been explored before, \cite{weinandeep} does not face the type of singularity issue arises in our application.

\section{Proposed method}
\label{section:architecture}
In this section, we present the general strategy of solving \eqref{BK} using a neural-network. For simplicity, we let $\Omega = \mathbb{R}^d$, $U(\x)$ be a confining potential that gives rise to an equilibrium measure $\mu(\x):=\exp(-\beta U(\x))/Z(\beta)$ normalized on the region $\Omega \backslash (A \cup B)$, where  $Z(\beta) := \int_{ \Omega\setminus A \cup B}  \exp(-\beta U(\x)) d\x$.  Instead of working with the strong form \eqref{BK}, we solve the variational problem
\begin{equation}
\label{variational form}
\underset{q}{\text{arginf}} \frac{1}{Z(\beta)}\int_{ \Omega\setminus A \cup B} \vert \nabla q(\x) \vert^2 \exp(-\beta U(\x)) d\x, \quad q(\x)\vert_{\partial A} = 0, \quad q(\x)\vert_{\partial B} = 1,\ \text{boundary condition on}\ \partial \Omega.
\end{equation}
To see the boundary conditions for $q$ on $\partial \Omega$, let
$q^*(\x)$ be the minimizer of \eqref{variational form} and
$q(\x,\lambda) = q^*(\x) + \lambda \eta(\x)$. Since $q^*(\x)$ is a
stationary point, for any $\eta(\x)$
\begin{eqnarray}
\label{weak to strong}
 0&=&\frac{1}{2}\frac{\partial }{\partial \lambda}\int_{\Omega\setminus A \cup B} \vert \nabla q(\x,\lambda) \vert^2 \exp(-\beta U(\x)) d\x \bigg\vert_{\lambda = 0}\cr
 &=&  \int_{ \Omega\setminus A \cup B}  \nabla q^*(\x) \cdot \nabla \eta(\x)  \exp(-\beta U(\x)) d\x \cr
 &=&    \int_{\Omega\setminus A \cup B}  \nabla\cdot( \nabla q^*(\x) \eta(\x) \exp(-\beta U(\x)) ) d\x- \int_{\Omega\setminus A \cup B}  \eta(\x)  \nabla\cdot ( \nabla q^*(\x) \exp(-\beta U(\x))) d\x\cr
 &=& - \int_{\Omega\setminus A \cup B} \eta(\x)  \nabla\cdot ( \nabla q^*(\x) \exp(-\beta U(\x))) d\x\cr
 &=& -\int_{\Omega\setminus A \cup B}  \eta(\x) ( \Delta q^*(\x) - \beta \nabla U(\x) \cdot \nabla q^*(\x)) \exp(-\beta U(\x) )d\x.
\end{eqnarray}
The third equality follows from $\eta(\x) = 0$ on $\partial A, \partial B$, and requiring
\begin{equation}
\label{boundary integral}
\int_{\Omega\setminus A\cup B }  \nabla \cdot( \nabla q^*(\x) \eta(\x) \exp(-\beta U(\x)) ) d\x= \int_{\partial \Omega} \nabla q^*(\x) \eta(\x) \exp(-\beta U(\x))  d\mathbf{s}=0
\end{equation}
via imposing suitable boundary condition on $\partial \Omega$. Here $\int_{\partial \Omega}  d\mathbf{s}$ stands for the surface integral. When the domain is unbounded as the considered case, the condition
\begin{equation}
\label{unbounded}
\int_{\partial B_R} \nabla q^*(\x)  \exp(-\beta U(\x))  d\mathbf{s}\rightarrow 0 \ \text{as}\ R \rightarrow \infty
\end{equation}
where $B_R$ denotes a ball with radius $R$, can ensure \eqref{boundary integral}. Notice that we simply need $\nabla q(\x)$ to have subexponential growth as $\vert \x \vert\rightarrow \infty$ for \eqref{unbounded} to hold, as long as $\exp(-\beta U(\x)) \leq \exp(-a \vert \x\vert)$ when $\vert \x \vert >R$ for some $R, a>0$. The last equality in \eqref{weak to strong} implies that a solution to \eqref{variational form} provides a solution to \eqref{BK}. 


As mentioned earlier, to cope with the high-dimensionality of $q(\x)$, the proposed method consists of parameterizing $q(\x)$ as an NN function $q_\theta(\x)$. Instead of \eqref{variational form}, we solve
\begin{gather}
\label{variational form2}
\underset{\theta \in \mathbb{R}^p}{\text{argmin}} \frac{1}{Z(\beta)}\int_{\Omega\setminus A \cup B} \vert \nabla q_\theta(\x) \vert^2 \exp(-\beta U(\x)) d\x + \rho \int_{\partial A} q_\theta(\x)^2 d\mu_{\partial A}(\x)+ \rho \int_{\partial B} (q_\theta(\x)-1)^2d\mu_{\partial B}(\x),
\end{gather}
where the boundary conditions are only enforced as soft-constraints (with hardness tuned by the choice of $\rho$). The first integral is then approximated via sampling according to the overdamped Langevin process \eqref{langevin}. To approximate the second and third integrals, for our problems there exist rather convenient scheme for drawing samples from $\mu_{\partial A}(\x)$, $\mu_{\partial B}(\x)$. The choice of the measures on the boundaries is based on the consideration of sampling convenience.  In our examples, we mainly work with regions $A$ and $B$ being balls, therefore the samples on $\partial A$ and $\partial B$ are drawn by normalizing and recentering normally distributed samples. Note that we can rewrite \eqref{variational form2} as a single expectation 
\begin{equation}
\label{variational form3}
\underset{\theta \in \mathbb{R}^p}{\text{argmin}}\; \mathbb{E}_{\nu}  \Bigl( \vert \nabla q_\theta(\x) \vert^2 \chi_{\Omega\setminus A \cup B}(\x)   +  \frac{\rho}{\alpha} q_\theta(\x)^2 \chi_{\partial A}(\x) + \frac{\rho}{\alpha} (q_\theta(\x)-1)^2 \chi_{\partial B}(\x) \Bigr),
\end{equation}
if we define a mixture measure 
\begin{equation*}
  \nu(\x) = \frac{1}{1 + 2\alpha} \Bigl( \frac{1}{Z(\beta)} e^{-\beta U(\mathbf{x})} 
  d \mathbf{x} + \alpha \mu_{\partial A}(\mathbf{x}) + \alpha \mu_{\partial B} (\mathbf{x}) \Bigr),
\end{equation*}
where $\chi_\Sigma(\x)$ is the characteristic function of region $\Sigma$, $\alpha$ is a parameter that controls the proportion between  the  sample size in $\partial A \cup \partial B$ with the sample size in $\Omega \backslash (A \cup B)$. 
This allows us to solve for \eqref{BK} as an optimization problem
\eqref{variational form3} over NN weights using stochastic gradient
type methods based on stochastic approximation of the
expectation. While this is in principle straightforward, challenges
arise due to the specific nature of the high dimensional Fokker-Planck
equation we aim to solve. We discuss those challenges below and then
the proposed neural-network design to overcome them.

\subsection{Challenge in high $T$ regime}
\begin{figure}
	\centering
	\includegraphics[width=0.5\textwidth]{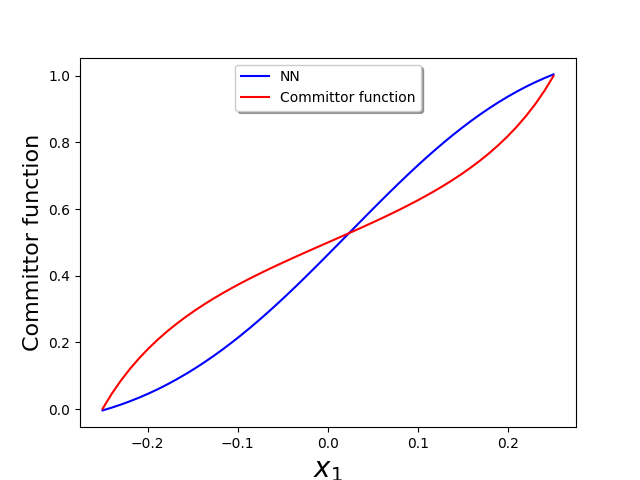}
	\caption{The solution to \eqref{2sphere} obtained from method of images (Red) and solving \eqref{variational form3} (Blue) without taking care of the singularity issue. }\label{figure:laplace}
\end{figure}

Although this method seems straight-forward, the main difficulty in using an NN to approximate the committor function is that in some situations, singularities are present within region $A$ and $B$. Consider the case where $A$ and $B$ are two balls of radius $r$ and $\Omega = \mathbb{R}^d, d\geq 3$, centered at $(-w_0/2,0,\ldots,0), (w_0/2,0,\ldots,0)$. When $T\rightarrow \infty $ and $\beta \rightarrow 0$, \eqref{BK} becomes
\begin{equation}
\label{2sphere}
\Delta q(\mathbf{x})=0\ \ \text{in}\ \Omega \setminus A\cup B, \quad q(\mathbf{x})\vert_{\partial A} = 0, \quad q(\mathbf{x})\vert_{ \partial B} = 1,\quad \nabla q(\x) \sim o(1/\vert \x \vert^{d-1}) \ \text{as}\ \vert \x \vert \rightarrow \infty.
\end{equation}
The last boundary condition is there in order to satisfy \eqref{unbounded}. A solution to \eqref{2sphere} can be obtained by first solving the Laplace's equation with Dirichlet's boundary conditions
\begin{equation}
\label{laplace}
\Delta \tilde q(\mathbf{x})=0\ \ \text{in}\ \Omega \setminus A\cup B, \quad \tilde q(\mathbf{x})\vert_{\partial A} = -1, \quad \tilde q(\mathbf{x})\vert_{ \partial B} = 1
\end{equation}
and letting $q(\x) = (1/2)(\tilde q(\x) + 1)$. A classical way to solve \eqref{laplace} analytically is via method of images. Using the Green's function 
\begin{equation}
G(\mathbf{x},\mathbf{y}) = \frac{\Gamma(d/2)}{(2\pi)^{d/2}\vert \mathbf{x} - \mathbf{y}\vert^{d-2}},
\end{equation}
which solves
\begin{equation}
\Delta G(\mathbf{x},\mathbf{y}) = \delta(\mathbf{x}-\mathbf{y})
\end{equation}
where $\Gamma(\cdot)$ is the gamma function, the solution of \eqref{laplace} can be obtained as
\begin{equation}
\label{images}
\tilde q(\mathbf{x}) = \sum_{i=0}^\infty c_i G( \mathbf{x} , \mathbf{y}^{+,i})  - \sum_{i=0}^\infty c_iG( \mathbf{x} , \mathbf{y}^{-,i})
\end{equation}
where
\begin{equation}
c_i = c_{i-1}(r/w_i)^{d-2},\quad  w_i =w_{0}-r^2/w_{i-1},\quad\mathbf{y}^{+,i} =  (w_i-w_0/2)\mathbf{e}_1,\quad \mathbf{y}^{-,i} = - (w_i-w_0/2)\mathbf{e}_1,
\end{equation}
where $c_0 = (2\pi)^{d/2}/\Gamma(d/2)$. Due to the singularities in $\tilde q$, the committor function may be steep near regions $A, B$ which can present difficulties when using an NN approximation. To illustrate, we let $d=3$, $w_0 = 1, r=0.15$ and we plot the solution of \eqref{2sphere} along the $x_1$-dimension in Fig. \ref{figure:laplace}. As a contrast, we minimize \eqref{variational form3} using an NN with 3 hidden-layers  and $\tanh$ nonlinearities, where each hidden-layer has 12 nodes. 3e+04 samples sampled uniformly from the box $[-2,2]^3$ are used in the optimization problem. We let $\rho=666$ and $\alpha = 1/15$ to enforce the boundary condition on $\partial A$ and $\partial B$. As shown in Fig. \ref{figure:laplace}, the NN has difficulty capturing the behavior of the committor function near the singularities.

\subsection{Challenge in low $T$ regime}
\begin{figure}
	\centering
	\includegraphics[width=0.5\textwidth]{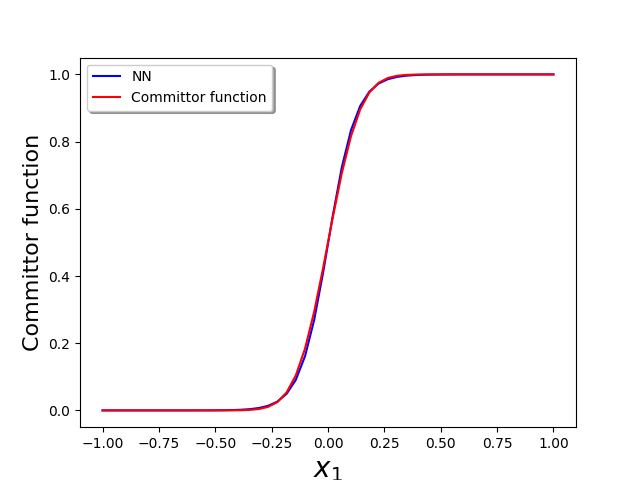}
	\caption{The committor function in \eqref{Wwell} along $x_1$ dimension when $T=0.05$ for an arbitrarily chosen $(x_2,\ldots,x_d)$ with $d=10$.}\label{figure:Wwell_smallT}
\end{figure}

A different type of singularities can exist in the low temperature regime. Consider the potential
\begin{equation}
\label{Wwell}
U(\x) = (x_1^2-1)^2 + 0.3\sum_{i=2}^d x_i^2,
\end{equation}
and
\begin{equation}
\label{Wwell AB}
A = \{x\in \mathbb{R}^d \vert x_1\leq -1\},\quad B = \{x\in \mathbb{R}^d \vert x_1\geq 1\}.
\end{equation}
Here, \eqref{Wwell} resembles a double potential well, and $\partial A$ and $\partial B$ are located in the potential wells. When the temperature is low, the equilibrium distribution for such $U(\x)$ is concentrated in $A$ and $B$ while the midpoint of $A$ and $B$ has a low density. Therefore, as $q(\x)$ goes from 0 to 1 when $x_1$ goes from -1 to 1, it is preferrable for $\vert \nabla q(\x) \vert$ to concentrate around the midpoint of $A$ and $B$ in order to have a low cost 
\begin{equation}
\frac{1}{Z(\beta)}\int_{\Omega\setminus A \cup B} \vert \nabla q_\theta(\x) \vert^2 \exp(-\beta U(\x)) d\x.
\end{equation} 
As an example, we plot the committor function when $T=0.05$ in Fig. \ref{figure:Wwell_smallT} when $d=10$. While the committor function is steep around $x_1=0$, unlike the case of high $T$, an NN with a single hidden layer with $\tanh$ activation function gives a good approximation. Since our goal is simply to show qualitatively that an NN is capable of handling such singularity issue, we defer the implementation details to Section \ref{section:numerical}.

\subsection{Neural-network architecture}
In this subsection, we present an NN network architecture that can deal with the aforementioned challenges. When solving problem \eqref{2sphere} via its variational formulation, a natural choice of the NN architecture is to mimic \eqref{images} and take
\begin{equation}
q_\theta(\x) = 1/2(\tilde q_\theta(\x) +1),\quad \tilde q_\theta(\x) := n_{\theta_1}(\x) G( \mathbf{x} , \mathbf{y}^{+,0}) + n_{\theta_2}(\x) G( \mathbf{x} , \mathbf{y}^{-,0})
\end{equation}
where $n_{\theta_1},n_{\theta_2}$ are neural-network parameterized functions, $\theta = [\theta_1^T,\theta_2^T ]^T$. By this ansatz, we explicitly remove the dominant singularities at $ \mathbf{y}^{+,0}, \mathbf{y}^{-,0}$. On the other hand, in order to determine the committor function of the transition process in the potential \eqref{Wwell} when $\beta \rightarrow \infty$, $(\tanh(wf(x_1))+1)/2$ where $w$ is a large scalar and  $f$ is certain smooth function may be used to capture the sharp transition around the midpoint between $\partial A$ and $\partial B$. This suggests using $\tanh$ as nonlinearity in an NN. 

\begin{figure}
	\centering
	\includegraphics[width=1\textwidth]{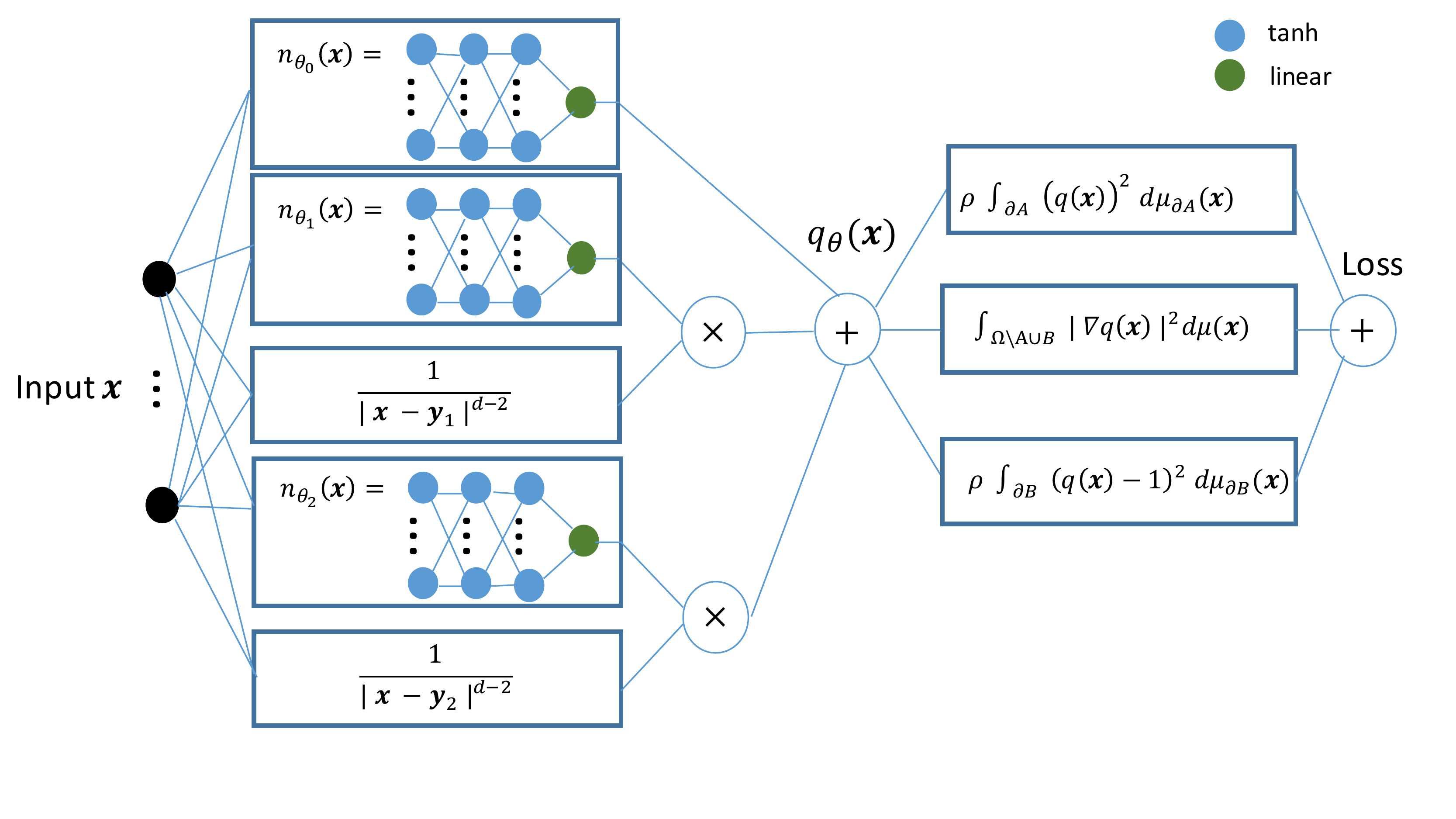}
	\caption{An example of the neural network architecture for a committor function with two $1/\vert \x\vert^{d-2}$ type singularities.  }\label{figure:architecture}
\end{figure}

\begin{figure}
	\centering
	\includegraphics[width=0.5\textwidth]{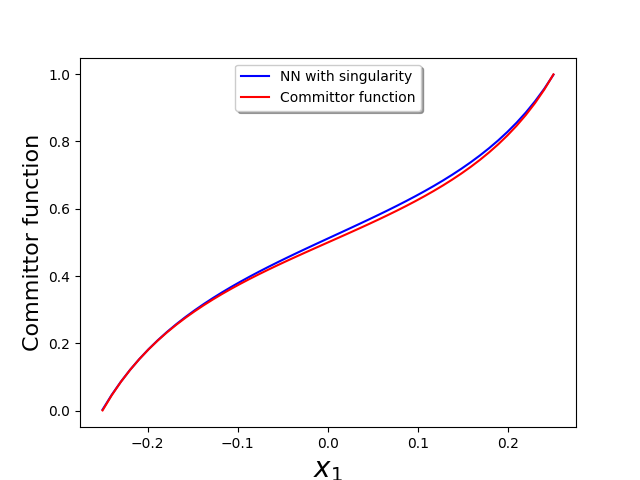}
	\caption{The solution to \eqref{BK} from method of images (Blue) and from solving \eqref{variational form3} when explicitly including $1/\vert \x \vert$ type singularities in the NN. }\label{figure:laplace with sing}
\end{figure}

Therefore,  to deal the issue of singularity, we propose the following neural-network architecture to solve for the committor function：
\begin{equation}
\label{ansatz}
 q_\theta(\x) := \sum_{k=1}^{N_s} n_{\theta_k}(\x) S_k( \mathbf{x} , \mathbf{y}^k)+ n_{\theta_0}(\x),
\end{equation}
where $n_{\theta_k}(\x)$'s are functions parameterized by the NN, $N_s$ is the number of singularities, and each $S_k(\x, \mathbf{y}^k)$ is a problem dependent function with singularity at $ \mathbf{y}^k$. The vector $\theta = [\theta_0^T,\ldots,\theta_{N_s}^T ]^T$ contains the parameters of the neural networks. Except $n_{\theta_0}(\x)$, each $n_{\theta_k}(\x)$ is an NN with 3 hidden layers where each hidden layer has 6 nodes. $n_{\theta_0}(\x)$ consists of multiple hidden layers each having 12 nodes. A hyperparameter we tune here is the number of hidden layers in $n_{\theta_0}(\x)$, where the choice of it is made using cross-validation. More precisely, the NN for committor function should give similar cost $\int_{\x\in \Omega\setminus A \cup B} \vert \nabla q_\theta(\x) \vert^2 \exp(-\beta U(\x)) d\x$ in training  and testing samples. We use $\tanh$ as the activation function of the hidden nodes. At high temperature, we expect  singularities of $1/\vert \x \vert ^{d-2}$ type to be dominant, whereas at low temperature the function $n_{\theta_0}(\x)$ with $\tanh$ nonlinearities should be the main contributor to the committor function. The pipeline of solving for $q_\theta(\x)$ is depicted in Fig. \ref{figure:architecture}. As shown in Fig. \ref{figure:laplace with sing}, when using such architecture to solve for the variational form of \eqref{2sphere}, we indeed recover the $1/\vert \x \vert$ type behavior near $A$ and $B$.

\section{Numerical experiments}
\label{section:numerical}

In this section, we evaluate the proposed method in a few numerical examples. In these examples, the minimization in \eqref{variational form3} using such NN architecture is done using the Adam \cite{kingma2014adam} optimizer, a variant of stochastic gradient descent, in the TensorFlow \cite{abadi2016tensorflow} engine.  The ratio $2\alpha$ of samples on $\partial A\cup \partial B$ to samples on $\Omega \setminus \partial A \cup \partial B$ are kept between 1/10 to 1/100. Then $\rho$ is tuned in order to have the boundary conditions satisfied with $10^{-3}$ accuracy. In all of the experiments, 2000 boundary samples are used, and we set the batch size to be 3000 in the Adam optimizer. We evaluate the performance using the following metric
\begin{equation}
E_1 = \frac{\| q_\theta - q \|_{L_2(\mu)}}{\| q\|_{L_2(\mu)}}, \quad E_2 = \frac{\vert v_R(q_\theta) - v_R(q) \vert}{v_R(q)}
\end{equation}
where $v_R(q) : = k_B T/Z(\beta) \int_{\Omega\setminus A \cup B} \vert \nabla q(x) \vert^2 \exp(-\beta U(x)) d\x$ is the rate of reaction. Note that $v_R$ is the energy one minimizes for $q$ in the variational formulation. To calculate these errors, we generate samples by simulating the stochastic process \eqref{langevin}. 

In the first numerical experiment, we solve for the committor function in the potential well \eqref{Wwell} with regions $A$ and $B$ being \eqref{Wwell AB} when $d=10$. In this case, $q(\x) = f(x_1)$ where
\begin{equation}
\frac{d^2 f(x_1)}{d x_1^2} - 4 x_1 (x_1^2-1) \frac{d f(x_1)}{d x_1} = 0,\quad f(-1)=0, \quad f(1)=1.
\end{equation}
To solve this problem using an NN, we set $N_s=0$ in \eqref{ansatz} as there is no singularity in this problem. In $n_{\theta_0}(\x)$, only one hidden layer is used. In this example, we sample differently from what is presented in \eqref{variational form3}.  When $T$ is small, it is difficult to obtain sufficient samples near the $\x=0$ saddle point of $U(\x)$. Therefore, instead of working with \eqref{variational form3} directly, we sample $x_1$ uniformly from $[-1,1]$, $(x_2,\ldots,x_d)$ from a $d-1$-dimensional gaussian distribution, and change the first term of the integrand in \eqref{variational form3} from  $ \vert \nabla q_\theta(\x) \vert^2 \chi_{\Omega\setminus A \cup B}(\x)$ to 
\begin{equation}
\frac{1}{ \int_{[-1,1]}\exp(-\beta (x_1^2-1)^2) dx_1 }\vert \nabla q_\theta(\x) \vert^2  \exp(-\beta (x_1^2-1)^2) \chi_{\Omega\setminus A \cup B}(\x)
\end{equation}
to ensure sufficient sample coverages along $x_1$. In this case, 
\begin{equation}
\nu(\x) =   \frac{1}{1 + 2\alpha} \Bigl( \frac{1}{2(2\pi T/0.6)^{(d-1)/2}}\chi_{[-1,1]}(x_1)\exp(-0.3\beta \sum_{i=2}^d x_i^2)
d \mathbf{x} + \alpha \mu_{\partial A}(\mathbf{x}) + \alpha \mu_{\partial B} (\mathbf{x}) \Bigr).
\end{equation}
For a subset of these samples, we let $x_1 = 1,-1$ to get the samples on the boundaries $\partial A, \partial B$. We use a separate batch of samples, serving as validation dataset, to determine $E_1$ and $E_2$. In Table \ref{table:Wwell} we report the error and the number of samples used for solving this problem in dimension $d=10$ with temperature $T=0.2, 0.05$. 

\begin{table}[ht]
	\centering 
	\begin{tabular}{ c c c c c c c c } 
		\hline\hline 
		 $T$ & $E_1$ & $E_2$ &  \begin{tabular}{@{}c@{}} No. of\\ parameters \end{tabular}& $\rho$ & \begin{tabular}{@{}c@{}}No. of \\ samples in \\ $\Omega\setminus  A \cup B$ \end{tabular} & $\alpha$ & \begin{tabular}{@{}c@{}}No. of \\ testing samples\end{tabular}  \\ [0.5ex] %
		\hline\hline 
		 0.2 & 0.0054 & 0.0063 & 145 & 50 & 2.0e+04 & 1/20  & 1e+05\\
		 0.05 &  0.012 & 0.020  & 145 & 0.5 & 2.0e+04 & 1/20 & 1e+05 \\
		\hline 
	\end{tabular}
	\caption{Results for the double well potential \eqref{Wwell} between two planes.}\label{table:Wwell} 
\end{table}

\begin{figure}
	\centering
	
	\subfloat[Including singularity.]{\includegraphics[width=0.4\textwidth]{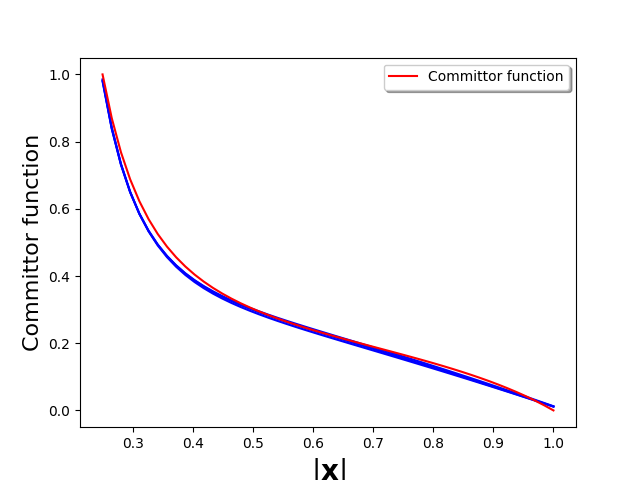}}
	\subfloat[Without singularity.]{\includegraphics[width=0.4\textwidth]{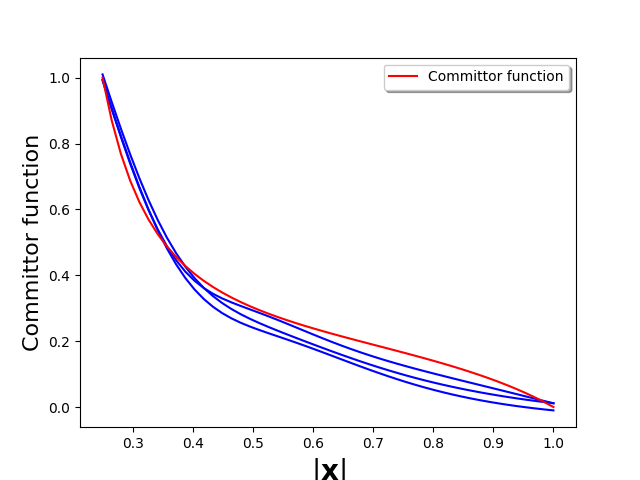}}
	\caption{The committor function for the stochastic process \eqref{ON sphere} between a pair of coecentric spheres as a function of $\vert \x \vert$. We compare the case when singularity is explictly included in the NN parameterization and the case when singularity is not included (in (a) and (b) respectively). Red: The ground truth committor function. Blue: The NN parameterized committor function along three different choices of radial direction.  }\label{figure:coecentric}
\end{figure}

In the second experiment, we solve for the committor function for the transition process between a pair of coecentric spheres, with potential
\begin{equation}
\label{ON sphere}
U(\x) = 10 \vert \x\vert^2
\end{equation}
and the regions
\begin{equation}
A = \{\x \in \mathbb{R}^d \vert \ \vert \x\vert\geq a\},\quad B = \{\x\in \mathbb{R}^d \vert  \ \vert \x\vert\leq b\}.
\end{equation}
In this example, even with moderate $T$, the committor function still display a singular behavior $q\sim1/\vert \x\vert^{d-2}, d\geq 3$. Therefore in \eqref{ansatz} we let $N_s = 1$, $S_1 = 1/\vert \x\vert^{d-2}$. We use 3 hidden layers for $n_{\theta_0}$. The equilibrium density is proportional to $\exp(-\beta \vert \x\vert^2)$, therefore the samples can be drawn from the gaussian distribution. The samples on the two boundaries are obtained via rescaling samples from the normal distribution to have norm $a$ or $b$. The results for $T = 2$, $d=6$, $a= 1$, $b=0.25$ are summarized in Table \ref{table:coecentric}. We compare the solution with and without including the $1/\vert \x\vert^{d-2}$ type singularity. It is worth noting that the explicit inclusion of singularity is rather important for this example even at moderate temperature.  In Fig. \ref{figure:coecentric}a, we plot $q_\theta(\x)$ along several randomly chosen radial directions to check whether $q_\theta(\x)$ is close to a single-variable function when explicitly including a singular function in the NN architecture. In Fig. \ref{figure:coecentric}b, we plot the NN committor function when the singularity is not explicitly taken care of.

\begin{table}[ht]
	\centering 
	\begin{tabular}{ c c c c c c c c } 
		\hline\hline 
		$N_s$ & $E_1$ & $E_2$ &  \begin{tabular}{@{}c@{}} No. of\\ parameters \end{tabular} &$\rho$ & \begin{tabular}{@{}c@{}}No. of \\ samples in \\$\Omega\setminus  A \cup B$ \end{tabular} & $\alpha$ & \begin{tabular}{@{}c@{}}No. of \\ testing samples\end{tabular} \\ [0.5ex] %
		\hline\hline 
		 1 & 0.053 & 0.015 & 542 & 5.3e+02 & 3e+04 & 1/30  & 1e+05\\
		 0 & 0.17 & 0.078 & 542 & 5.3e+02 & 3e+04 & 1/30  & 1e+05\\ [1ex] 
		\hline 
	\end{tabular}
	\caption{Results for the coecentric spheres example. We compare the cases when $N_s = 0$ and $N_s=1$.}\label{table:coecentric} 
\end{table}

\begin{figure}[!ht]
	\centering
	\subfloat[$T=40$ equilibrium distribution.]{\includegraphics[width=0.35\columnwidth]{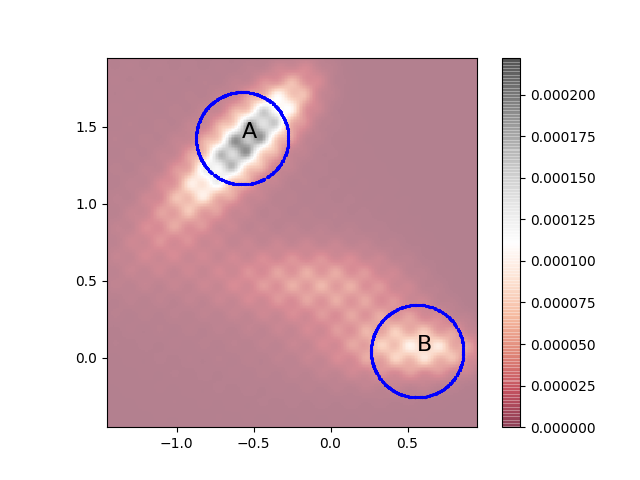}}
	\subfloat[$T=40$ committor function]{\includegraphics[width=0.35\columnwidth]{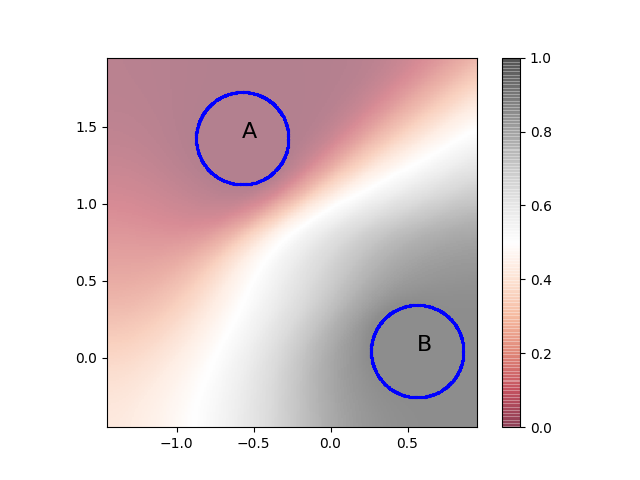}}
	\subfloat[$T=40$ NN committor function]{\includegraphics[width=0.35\columnwidth]{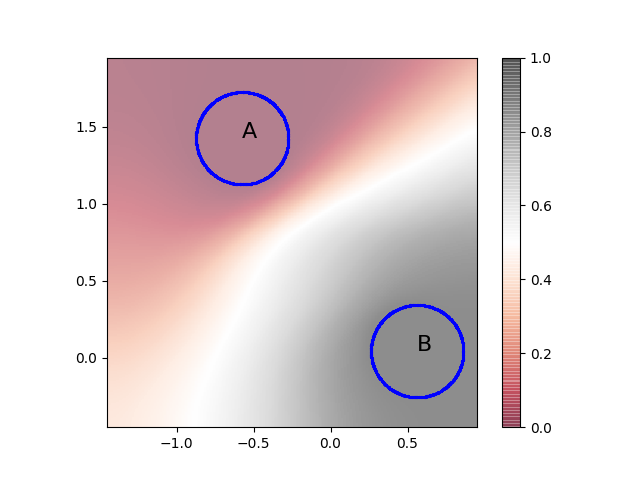}}
	\qquad
	\subfloat[$T=22$ equilibrium distribution.]{\includegraphics[width=0.35\columnwidth]{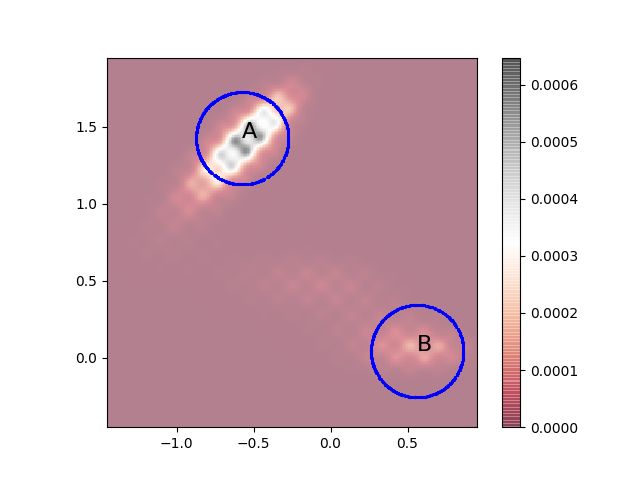}}
	\subfloat[$T=22$ committor function]{\includegraphics[width=0.35\columnwidth]{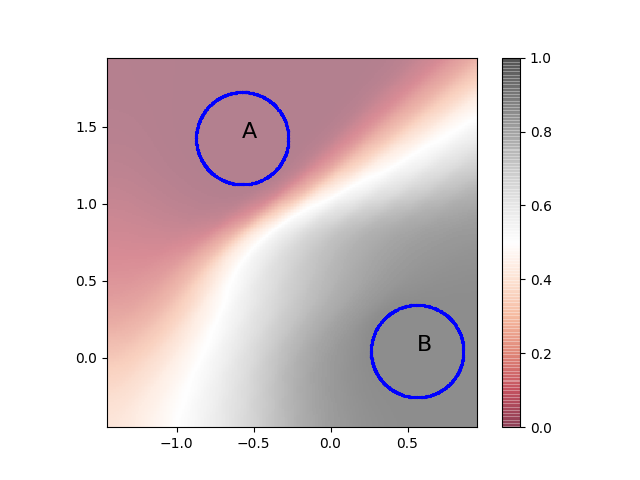}}
	\subfloat[$T=22$ NN committor function]{\includegraphics[width=0.35\columnwidth]{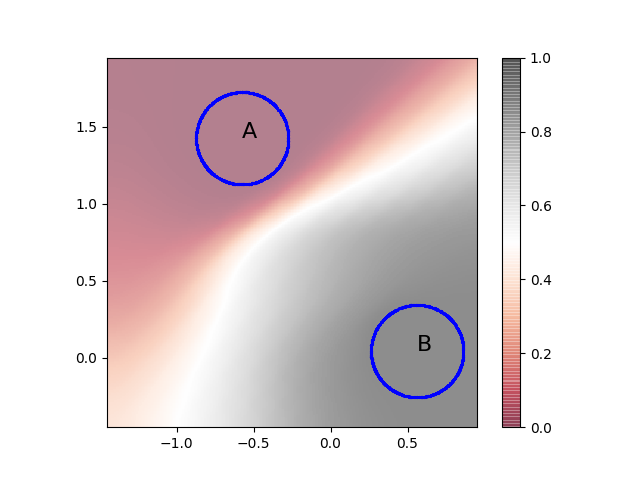}}
	\caption{Figures for the rugged Muller potential on a 2-dimensional hyperplane. (a) and (d): The equilibrium distribution when $T=40,22$ for the rugged-Muller potential. (b)  and (c): The ground truth committor function and the NN parameterized committor function for $T=40$.  (e) and (f): The ground truth committor function and the NN parameterized committor function for $T=22$. }\label{figure:RM}
	
\end{figure}

In the third experiment, we work with the rugged-Muller potential 
\begin{multline}
\label{RM potential}
U(x,y) = \sum_{i=1}^4 D_i \exp(a_i(x_1-X_i)^2+b_i(x_1-X_i)(x_2-Y_i)+c_i(x_2-Y_i)^2) + \gamma \sin(2k\pi x)\sin(2k\pi y) + \frac{1}{2\sigma^2} \sum_{i=3}^d x_i^2
\end{multline}
considered in \cite{lai2017point}. This is a Muller potential perturbed by a rugged potential in the first two dimension, where the roughness is controlled by $\gamma,k$. In the rest of the dimensions, we place a quadratic potential well where its strength is controlled by $\sigma$. The domain $\Omega = [-1.5,1]\times [-0.5,2]$. The parameters in \eqref{RM potential} are taken from \cite{lai2017point}, for completeness we provide them in the following:
\begin{eqnarray}
[a_1,a_2,a_3,a_4]        &=&   [-1, -1, -6.5, 0.7],\cr
[b_1, b_2, b_3, b_4] &=& [0, 0, 11, 0.6],\cr
[c_1, c_2, c_3, c_4]  &=& [-10, -10, -6.5, 0.7],\cr
[D_1,D_2,D_3,D_4] &=& [-200, -100, -170, 15],\cr
[X_1, X_2, X_3, X_4] &=& [1, 0, -0.5, -1],\cr
[Y_1, Y_2, Y_3, Y_4] &=& [0, 0.5, 1.5, 1].
\end{eqnarray}
In this example, we let $T=40,22$, regions $A$ and $B$ being two balls with radius 0.1 centered at $(-0.57,  1.43)$ and $(-0.56,0.044)$. The points are again sampled using Euler-Maruyama scheme. The ground truth is obtained via applying finite element method on uniform grid to \eqref{BK}, where the code is provided by the authors of \cite{lai2017point}. We use an NN with two singularities of type $\log(\vert \x - \mathbf{y}\vert)$ where $\mathbf{y}$ is the position of singularity, and $n_{\theta_0}$ that has 3 hidden layers. The results are reported in Table \ref{table:muller} and the contours of the committor function are shown in Fig. \ref{figure:RM}. As shown in the table, although we can achieve few percents accuracy, for the case with lower temperature more samples are needed to determine the committor function (since the equilibrium distribution is less smooth). 
\begin{table}[ht]
	\centering 
	\begin{tabular}{c c c c c c c c c } 
		\hline\hline 
		$(T,\sigma)$ & $E_1$ & $E_2$ &\begin{tabular}{@{}c@{}} No. of\\ parameters \end{tabular}  &$\rho$ & \begin{tabular}{@{}c@{}}No. of \\ samples in\\ $\Omega\setminus  A \cup B$ \end{tabular} & $\alpha$ & \begin{tabular}{@{}c@{}}No. of \\ testing samples\end{tabular} \\ [0.5ex] %
		\hline\hline 
		(40, 0.05) & 0.057 & 0.035 & 1011 &3.8e+02 & 7.4e+04 & 1/74   & 7.4e+04\\
		(22, 0.05) & 0.037 & 0.036 & 1011 &1.3e+02 & 1.5e+05 & 1/150  & 1.5e+05\\
		\hline 
	\end{tabular}
	\caption{Results for the rugged Muller potential example. }\label{table:muller} 
\end{table}

\section{Conclusion}
\label{section:conclusion}
In this note, we develop method based on neural-network to represent
the high-dimensional committor function. The neural-network parameters
are found via optimizing the variational form of the Fokker-Planck
equation.  In order to better approximate the committor function, the
NN function has to be designed carefully in order to deal with the
singularities in high and low $T$ regime. Through numerical
experiments, we show the usefulness of the proposed alternative
approach in dealing with high-dimensional partial differential
equations. We remark that the quality of the learned committor
function depends crucially on sampling. When the temperature is low,
due to the sparsity of samples between regions $A$ and $B$ when a
naive sampling scheme is used, the NN approximation to the committor
function tends to make a transition that is too sharp compare to the
ground truth. The usage of enhanced sampling schemes, for example
using the currently learned NN to guide further sampling, is certainly
an important future direction to investigate.



\bibliographystyle{abbrv}
\bibliography{bibref}

\begin{thebibliography}{10}

\bibitem{abadi2016tensorflow}
M.~Abadi, A.~Agarwal, P.~Barham, E.~Brevdo, Z.~Chen, C.~Citro, G.~S. Corrado,
  A.~Davis, J.~Dean, M.~Devin, S.~Ghemawat, I.~Goodfellow, A.~Harp, G.~Irving,
  M.~Isard, Y.~Jia, R.~Jozefowicz, L.~Kaiser, J.~Kudlur, Manjunath~Levenberg,
  D.~Mane, R.~Monga, S.~Moore, D.~Murray, C.~Olah, M.~Schuster, J.~Shlens,
  B.~Steiner, I.~Sutskever, K.~Talwar, P.~Tucker, V.~Vanhoucke, V.~Vasudevan,
  F.~Viegas, O.~Vinyals, P.~Warden, M.~Wattenberg, M.~Wicke, Y.~Yu, and
  X.~Zheng.
\newblock Tensorflow: Large-scale machine learning on heterogeneous distributed
  systems.
\newblock {\em arXiv preprint arXiv:1603.04467}, 2016.

\bibitem{berg2017unified}
J.~Berg and K.~Nystr{\"o}m.
\newblock A unified deep artificial neural network approach to partial
  differential equations in complex geometries.
\newblock {\em arXiv preprint arXiv:1711.06464}, 2017.

\bibitem{carleo2017solving}
G.~Carleo and M.~Troyer.
\newblock Solving the quantum many-body problem with artificial neural
  networks.
\newblock {\em Science}, 355(6325):602--606, 2017.

\bibitem{coifman2008diffusion}
R.~R. Coifman, I.~G. Kevrekidis, S.~Lafon, M.~Maggioni, and B.~Nadler.
\newblock Diffusion maps, reduction coordinates, and low dimensional
  representation of stochastic systems.
\newblock {\em Multiscale Modeling \& Simulation}, 7(2):842--864, 2008.

\bibitem{coifman2006diffusion}
R.~R. Coifman and S.~Lafon.
\newblock Diffusion maps.
\newblock {\em Applied and computational harmonic analysis}, 21(1):5--30, 2006.

\bibitem{ERenVa:05}
W.~E, W.~Ren, and E.~Vanden-Eijnden.
\newblock Finite temparture string method for the study of rare events.
\newblock {\em J. Phys. Chem. B}, 109:6688--6693, 2005.

\bibitem{weinan2006towards}
W.~E and E.~Vanden-Eijnden.
\newblock Towards a theory of transition paths.
\newblock {\em Journal of statistical physics}, 123(3):503, 2006.

\bibitem{EVa:10}
W.~E and E.~Vanden-Eijnden.
\newblock Transition path theory and path-finding algorithms for the study of
  rare events.
\newblock {\em Ann. Rev. Phys. Chem.}, 61:391--420, 2010.

\bibitem{weinandeep}
W.~E and B.~Yu.
\newblock The deep {R}itz method: A deep learning-based numerical algorithm for
  solving variational problems.
\newblock {\em Communications in Mathematics and Statistics}, pages 1--12.

\bibitem{hinton2006reducing}
G.~E. Hinton and R.~R. Salakhutdinov.
\newblock Reducing the dimensionality of data with neural networks.
\newblock {\em Science}, 313(5786):504--507, 2006.

\bibitem{kingma2014adam}
D.~Kingma and J.~Ba.
\newblock Adam: A method for stochastic optimization.
\newblock {\em arXiv preprint arXiv:1412.6980}, 2014.

\bibitem{lagaris1998artificial}
I.~E. Lagaris, A.~Likas, and D.~I. Fotiadis.
\newblock Artificial neural networks for solving ordinary and partial
  differential equations.
\newblock {\em IEEE Transactions on Neural Networks}, 9(5):987--1000, 1998.

\bibitem{lai2017point}
R.~Lai and J.~Lu.
\newblock Point cloud discretization of {F}okker-{P}lanck operators for
  committor functions.
\newblock {\em Multiscale Model. Simul. in press; arXiv preprint
  arXiv:1703.09359}, 2017.

\bibitem{lecun2015deep}
Y.~LeCun, Y.~Bengio, and G.~Hinton.
\newblock Deep learning.
\newblock {\em Nature}, 521(7553):436--444, 2015.

\bibitem{LuNolen:15}
J.~Lu and J.~Nolen.
\newblock Reactive trajectories and the transition path process.
\newblock {\em Probab. Theory Related Fields}, 161:195--244, 2015.

\bibitem{schmidhuber2015deep}
J.~Schmidhuber.
\newblock Deep learning in neural networks: An overview.
\newblock {\em Neural networks}, 61:85--117, 2015.

\bibitem{sirignano2017dgm}
J.~Sirignano and K.~Spiliopoulos.
\newblock {DGM: A} deep learning algorithm for solving partial differential
  equations.
\newblock {\em arXiv preprint arXiv:1708.07469}, 2017.

\bibitem{VaVe:09}
E.~Vanden-Eijnden and M.~Venturoli.
\newblock Revisiting the finite temperature string method for the calculation
  of reaction tubes and free energies.
\newblock {\em J. Chem. Phys.}, 130:194103, 2009.

\bibitem{zhang2017reinforced}
L.~Zhang, H.~Wang, and W.~E.
\newblock Reinforced dynamics of large atomic and molecular systems.
\newblock {\em arXiv preprint arXiv:1712.03461}, 2017.

\end{thebibliography}

\end{document}